\documentclass{article}

\usepackage[english]{babel}
\usepackage[square,sort,comma,numbers]{natbib}

\usepackage[final]{nips_2016} 
\usepackage[utf8]{inputenc} 
\usepackage[T1]{fontenc}    
\usepackage{url}            
\usepackage{booktabs}       
\usepackage{amsfonts}       
\usepackage{nicefrac}       
\usepackage{microtype}      
\usepackage{graphicx}
\graphicspath{{figures/}}
\usepackage{subcaption}
\usepackage{amsmath}

\title{Active One-shot Learning}

\author{
  Mark Woodward \\
  Independent Researcher \\
  \texttt{mwoodward@cs.stanford.edu} \\
  \And
  Chelsea Finn \\
  Berkeley AI Research (BAIR) \\
  \texttt{cbfinn@eecs.berkeley.edu} \\
}

\begin{document}

\maketitle

%
\begin{abstract}
Recent advances in one-shot learning have produced models that can learn from a handful of labeled examples, for passive classification and regression tasks.
%
This paper combines reinforcement learning with one-shot learning, allowing the model to decide, during classification, which examples are worth labeling.
%
We introduce a classification task in which a stream of images are presented and, on each time step, a decision must be made to either predict a label or pay to receive the correct label. %
We present a recurrent neural network based action-value function, and demonstrate its ability to learn how and when to request labels.
Through the choice of reward function, the model can achieve a higher prediction accuracy than a similar model on a purely supervised task, or trade prediction accuracy for fewer label requests.
\end{abstract}

\section{Introduction} \label{sec:introduction}

Active learning, a special case of semi-supervised learning, is an approach for reducing the amount of supervision needed for performing a task, by having the model select which datapoints should be labeled. Active learning is particularly useful in domains where there is some large cost associated with making a mistake and another, typically smaller, cost associated with requesting a label. Applications have included domains where human or computer safety is at risk, such as cancer classification~\cite{liu2004active}, malware detection~\cite{moskovitch2008active}, and autonomous driving~\cite{laskey2016shiv,zhang2016query}. Methods for active learning involve strategies such as selecting the data points for which the model is the most uncertain, or picking examples which are expected to be the most informative. However, these methods often use heuristics for handling the computational complexity of approximating the risk associated with selection. Can we instead replace these heuristics with learning?

In this work, we combine meta-learning with reinforcement learning to \emph{learn} an active learner. In particular, we consider the online setting of active learning, where an agent is presented with examples in a sequence, and must choose whether to label the example or request the true label. 
Extending the recently proposed one-shot learning approach by Santoro et al.~\cite{santoro2016osl}, we develop a method for training a deep recurrent model to make labeling decisions.
%
Unlike prior one-shot learning approaches which use supervised learning, we treat the model as a policy with actions that include labeling and requesting a label, and train the policy with reinforcement learning. 
As a result, our trained model can make effective decisions with only a few labeled examples.

Our primary contribution is to present a method for learning an active learner using deep reinforcement learning. We evaluate our method on an Omniglot image classification task. Our preliminary results show that our proposed model can learn from only a handful of requested labels and can effectively trade off prediction accuracy with reduced label requests via the choice of reward function.
To the best of our knowledge, our model is the first application of reinforcement learning with deep recurrent models to the task of active learning.

\section{Related Work} \label{sec:related-work}




Active learning deals with the problem of choosing an example, or examples, to be labeled from a set of unlabeled examples~\cite{settles2009all}. 
We consider the setting of single pass active learning, in which a decision must be made on examples as they are pulled from a stream. Generally, methods for doing so have relied on heuristics such as similarity metrics between the current example and examples seen so far~\cite{lughofer2012spa}, or uncertainty measures in the label prediction~\cite{lughofer2012spa, chu2011uoa}. The premise of active learning is that there are costs associated with labeling and with making an incorrect prediction. Reinforcement learning allows for the explicit specification of those costs, and directly finds a labelling policy to optimize those costs. Thus, we believe that reinforcement learning is a natural fit for active learning.
We use a deep recurrent neural network function approximator for representing the action-value function.
While there have been numerous applications of deep neural networks to the related problem of semi-supervised learning~\cite{kingma2014ssl,rasmus2015ssl}, the application of deep learning to active learning problems has been limited~\cite{krogh1995nne}.

Our model is very closely related to recent approaches to meta-learning and one-shot learning. Meta-learning has been successfully applied to supervised learning tasks~\cite{santoro2016osl,vinyals2016mno}, with key insights being training on short episodes with few class examples and randomizing the labels and classes in the episode. We propose to combine such approaches for one-shot learning with reinforcement learning, to learn an agent that can make labelling decisions online. The task and model we propose is most similar to the model proposed by Santoro et al.~\cite{santoro2016osl}, in which the model must predict the label for a new image at each time step, with the true label received, as input, one time step later. We extend their task to the active learning domain by withholding the true label, unless the model requests it, and training the model with reinforcement learning, rewarding accurate predictions and penalizing incorrect predictions and label requests. Thus, the model must learn to consider its own uncertainty before making a prediction or requesting the true label.

\section{Preliminaries} \label{sec:preliminaries}

We will now briefly review reinforcement learning as a means of introducing notation. 
Reinforcement learning aims to learn a policy that maximizes the expected sum of discounted future rewards. Let $\pi(s_t)$ be a policy which takes a state, $s_t$, and outputs an action, $a_t$ at time $t$. One way to represent the optimal policy, $\pi^*(s_t)$, is as the action that maximizes the optimal action-value function, $Q^*(s_t, a_t)$, which specifies the expected sum of discounted future rewards for taking action $a_t$ in state $s_t$ and acting optimally from then on:

\begin{equation}
  a_t = \pi^*(s_t) = \mathop{{\arg\!\max}}_{a_t}Q^*(s_t, a_t).
\end{equation}

The following Bellman equation, for action-value functions, holds for the optimal $Q^*(s_t,a_t)$:

\begin{equation} \label{eqn:bellman}
  Q^*(s_t, a_t) = \mathbb{E}_{s_{t+1}} [r_t + \gamma \max_{a_{t+1}}Q^*(s_{t+1},a_{t+1})],
\end{equation}

where $r_t$ is the reward received after taking action $a_t$ in state $s_t$, and $\gamma$ is the discount factor for future rewards.


Many reinforcement learning algorithms use a function approximator for representing $Q(s_t, a_t)$ and optimize its parameters by minimizing the Bellman error, which can be derived from Equation~\ref{eqn:bellman} as the following:

\begin{equation}
  \mathcal{L}(\Theta) := \sum_t[Q_{\Theta}(o_t,a_t) - (r_t + \gamma \max_{a_{t+1}}Q_{\Theta}(o_{t+1}, a_{t+1}))]^2,
\end{equation}

where $\Theta$ are the parameters of the function approximator, 
and the algorithm receives observations $o_t$, such as images, rather than states $s_t$.
This is the equation that we optimize in the experiments below. We use a neural network to represent $Q$, which we optimize via stochastic gradient descent. Note that in the experiments we do not use a separate target network as in Mnih et al.~\cite{mnih2013pad}.

\section{Task Methodology} \label{sec:task-methodology}

\begin{figure}[t]
  \centering
  \begin{subfigure}{0.52\textwidth}
    \includegraphics[height=1.6in]{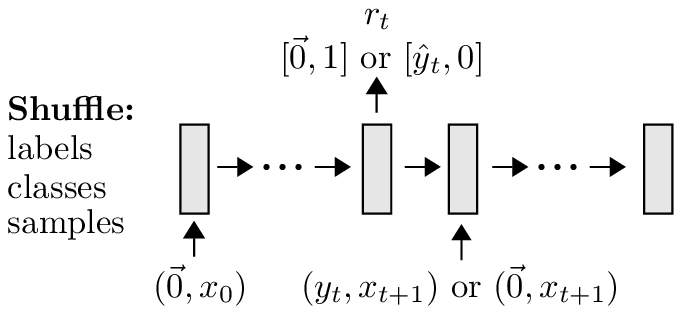}
    \caption{Task structure}
    \label{fig:task-a}
  \end{subfigure}
  { \unskip \vrule depth 0.5in }
  \begin{subfigure}{0.20\textwidth}
    \centering
    \includegraphics[height=1.6in]{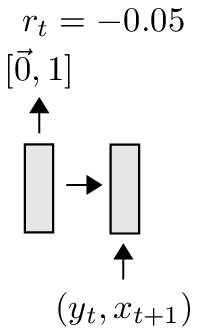}
    \caption{Request label}
    \label{fig:task-b}
  \end{subfigure}
  \begin{subfigure}{0.26\textwidth}
    \centering
    \includegraphics[height=1.6in]{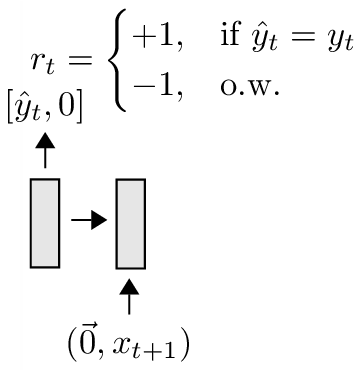}
    \caption{Predict label}
    \label{fig:task-c}
  \end{subfigure}
  \caption{Task structure. (a) An Omniglot image, $x_t$, is presented at each step of the episode. The model outputs a one-hot vector of length $c+1$, where $c$ is the number of classes per episode. (b) The model can request the label for $x_t$ by setting the final bit of the output vector. The reward, $r_t$, is $-0.05$ for a label request. The true label, $y_t$, for $x_t$ is then provided at the next time step along with $x_{t+1}$. (c) Alternatively, the model can make a prediction by setting one of the first $c$ bits of the output vector, designated $\hat{y}_t$. $r_t$ is $+1$ if the prediction is correct or $-1$ if not. If a prediction is made at time $t$, then no information about $y_t$ is supplied at time $t+1$. (a) For each episode, the classes to be presented, their labels, and the specific samples are all shuffled.}
  \label{fig:task}
\end{figure}


Similar to recent work on one-shot learning~\cite{santoro2016osl,vinyals2016mno}, we train with short episodes and a few examples per class, varying classes and randomizing labels between episodes. The intuition is that we want to delegate episode specific details to activations and fit the model weights to the general meta-task of learning to label.


Figure~\ref{fig:task} describes the meta-learning task addressed in this paper. The model receives an image, $x_t$, at each time step of the episode and may either predict the label of that image or request the label for that image. If the label is requested then the true label, $y_t$, is included along with the next image $x_{t+1}$ in the next observation $o_{t+1}$. The action, $a_t$, is a one-hot vector consisting of the optionally predicted label, $\hat{y}_t$, followed by a bit for requesting the label. Since only one bit can be set, the model can either make a label prediction or request the label. If a prediction is made, and thus no label is requested, a zero-vector, $\vec{0}$, is included in the next observation instead of the true label.

On each time step, one of three rewards is given: $R_{req}$, for requesting the label, $R_{cor}$, for correctly predicting the a label, or $R_{inc}$, for incorrectly predicting the label.

\begin{equation}
  r_t = 
  \begin{cases}
    R_{req}, & \text{if a label is requested} \\
    R_{cor}, & \text{if predicting and } \hat{y}_t = y_t \\
    R_{inc}, & \text{if predicting and } \hat{y}_t \neq y_t
  \end{cases}
\end{equation}

The objective is to maximize the sum of rewards received during the episode.

The optimal strategy involves maintaining a set of class representations and their corresponding labels, in memory. Then, upon receiving a new image $x_t$, the optimal strategy is to compare the representation for $x_t$ to the existing class representations, weighing the uncertainty of a match along with the cost of being incorrect, correct, or requesting a label, and either retrieving and outputting the stored label or requesting a new label. If the model believes $x_t$ to be an instance of a new class, then a class representation must be stored, the label must be requested, and the response must be stored and associated with the class representation.

\section{Reinforcement Learning Model} \label{sec:reinforcement-learning-model}

Our action-value function, $Q(o_t, a_t)$, is a long short-term memory (LSTM)~\cite{hochreiter1997lst}, connected to a linear output layer. $Q(o_t)$ outputs a vector, where each element corresponds to an action, similar to the DQN model~\cite{mnih2013pad}:

\begin{align}
    Q(o_t, a_t) &= Q(o_t) \cdot a_t \\
    Q(o_t) &= W^{hq}h_t + b^q
\end{align}

where $h_t$ is the output from the LSTM, $W^{hq}$ are the weights mapping from the LSTM output to action-values, and $b^q$ is the action-value bias. We use a basic LSTM with equations:

\begin{align}
    \hat{g}^f, \hat{g}^i, \hat{g}^o, \hat{c}_t &= W^{o}o_t + W^{h}h_{t-1} + b \\
    g^f &= \sigma(\hat{g}^f) \\
    g^i &= \sigma(\hat{g}^i) \\
    g^o &= \sigma(\hat{g}^o) \\
    c_t &= g^f\!\odot{}c_{t-1}+g^i\!\odot{}tanh(\hat{c}_t) \\
    h_t &= g^o\!\odot tanh(c_t)
\end{align}

where $\hat{g}^f$, $\hat{g}^i$, $\hat{g}^o$ are the forget gates, input gates, and output gates respectively, $\hat{c}_t$ is the candidate cell state, and $c_t$ is the new LSTM cell state. $W^o$ and $W^h$ are the weights mapping from the observation and hidden state, respectively, to the gates and candidate cell state, and $b$ is the bias vector. $\odot{}$ represents element-wise multiplication. $\sigma(\cdot)$ and $tanh(\cdot)$ are the sigmoid and hyperbolic tangent functions respectively.

\section{Experimental Results} \label{sec:experimental-results}

We evaluate our proposed one-shot learning model in an active-learning set-up for image classification. Our goal with the following experiments is to determine 1) whether or not the proposed model can learn, through reinforcement, how to label examples and when to instead request a label, and 2) whether or not the model is effectively reasoning about uncertainty when making its predictions.



\subsection{Setup}

We used the Omniglot dataset in all experiments~\cite{lake2015hlc}. Omniglot contains 1,623 classes of characters from 50 different alphabets, with 20 hand drawn examples per class, giving 32,460 total examples. We randomly split the classes into 1,200 training classes, and 423 test classes. Images were normalized to a pixel value between 0.0 and 1.0 and resized to 28x28 pixels.

Each episode consisted of 30 Omniglot images sampled randomly from 3 randomly sampled classes, without replacement. Note that the number of samples from each class may not have been equal. For each class in the episode, a random rotation in $\{0^\circ, 90^\circ, 180^\circ, 270^\circ\}$ was selected and applied to all samples from that class. The images were then flattened to 784 dimensional vectors, giving $x_t$. Each of the three sampled classes in the episode was randomly assigned a slot in a one-hot vector of length three, giving $y_t$. Each training step consisted of a batch of 50 episodes.

Unless otherwise specified, the rewards were: $R_{cor}=+1$, $R_{inc}=-1$, and $R_{req}=-0.05$. Epsilon-greedy exploration was used for actions selection during training, with $\epsilon=0.05$. If exploring, either the correct label, a random incorrect label, or the ``request label'' action was chosen, each with probability $1/3$. The discount factor, $\gamma$, was set to 0.5. We used an LSTM with 200 hidden units to represent $Q$. The weights of the model were trained using Adam with the default parameters~\cite{kingma2015ams}.

\subsection{Results}

\begin{figure}[t]
  \centering
  \begin{subfigure}{0.49\textwidth}
    \centering
    \includegraphics[width=1.0\columnwidth]{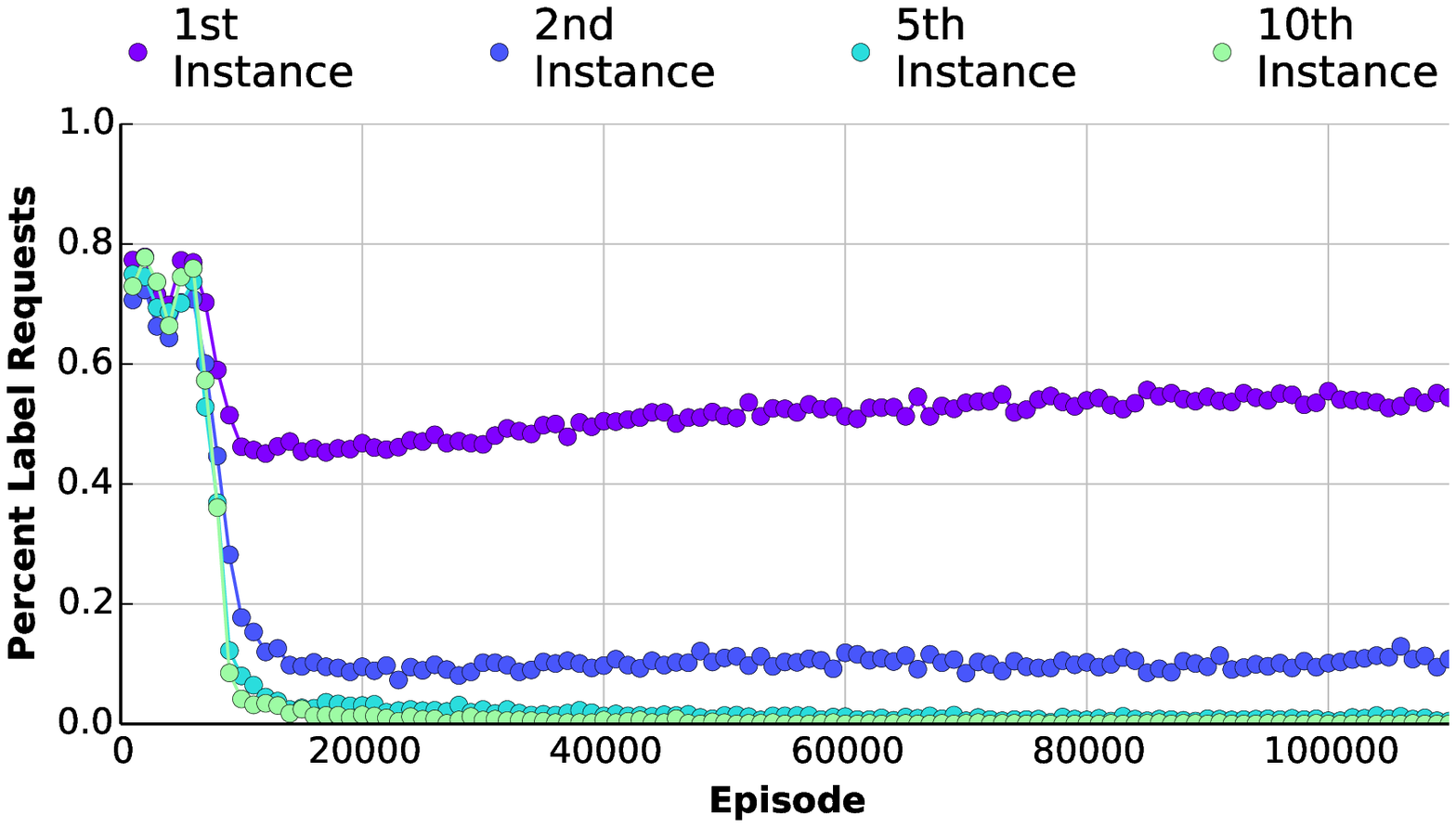}
    \caption{Label requests}
    \label{fig:accuracy-and-requests-requests-lstm}
  \end{subfigure}
  \begin{subfigure}{0.49\textwidth}
    \centering
    \includegraphics[width=1.0\columnwidth]{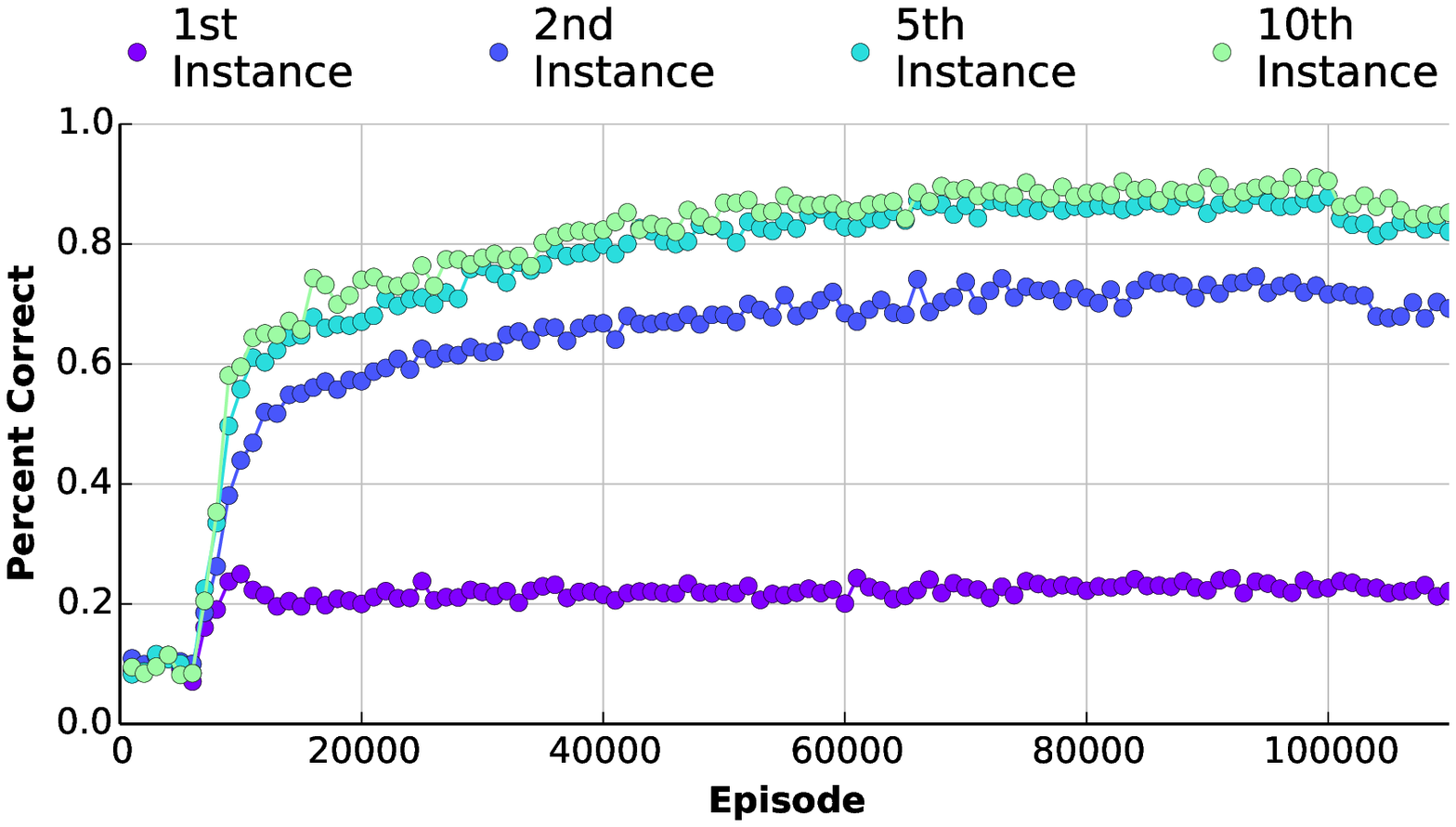}
    \caption{Accuracy}
    \label{fig:accuracy-and-requests-accuracy-lstm}
  \end{subfigure}
  \caption{Label requests (a) and accuracies (b) per episode batch for the 1\textsuperscript{st}, 2\textsuperscript{nd}, 5\textsuperscript{th}, and 10\textsuperscript{th} instances of all classes. The model requests fewer labels and has a higher accuracy on later instances of a class. At 100,000 episode batches, the training stops and the data switches to the test set.}
  \label{fig:accuracy-and-requests}
\end{figure}

We now present the results of our method.
During training, for each episode within a training batch, the time steps containing the 1\textsuperscript{st}, 2\textsuperscript{nd}, 5\textsuperscript{th}, and 10\textsuperscript{th} instances of all classes are identified. Figure~\ref{fig:accuracy-and-requests-requests-lstm} shows the percentage of label requests for each of these steps, as training progressed; whereas Figure~\ref{fig:accuracy-and-requests-accuracy-lstm} shows the percentage of actions corresponding to correct label predictions. Thus, we treat label requests as incorrect label predictions in this analysis. As seen in the plot, the model learns to make more label requests for early instances of a class, and fewer for later instances.
Correspondingly, the model is more accurate on later instances of a class. After the 100,000\textsuperscript{th} episode batch, training is ceased and the model is no longer updated. After 100,000 episode batches, 10,000 more episode batches were run on held-out classes from the test set. 

\begin{figure}[t]
  \centering
  \begin{subfigure}{0.34\textwidth}
    \centering
    \includegraphics[height=1.05in]{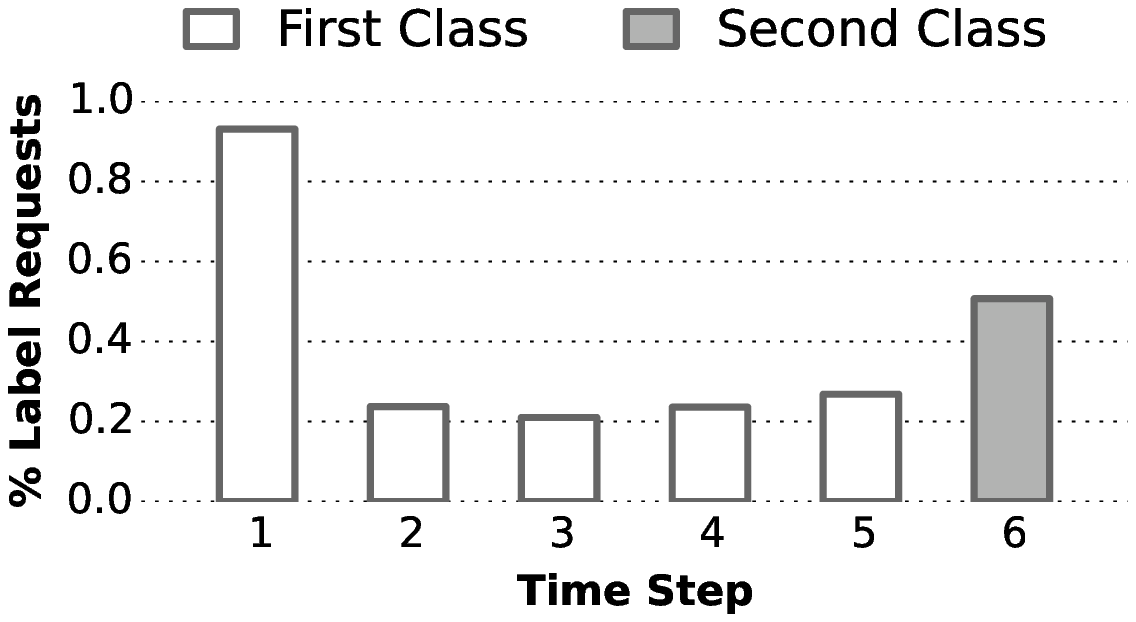}
    \caption{Switch classes after 5}
    \label{fig:class-switch-5}
  \end{subfigure}
  \begin{subfigure}{0.64\textwidth}
    \centering
    \includegraphics[height=1.05in]{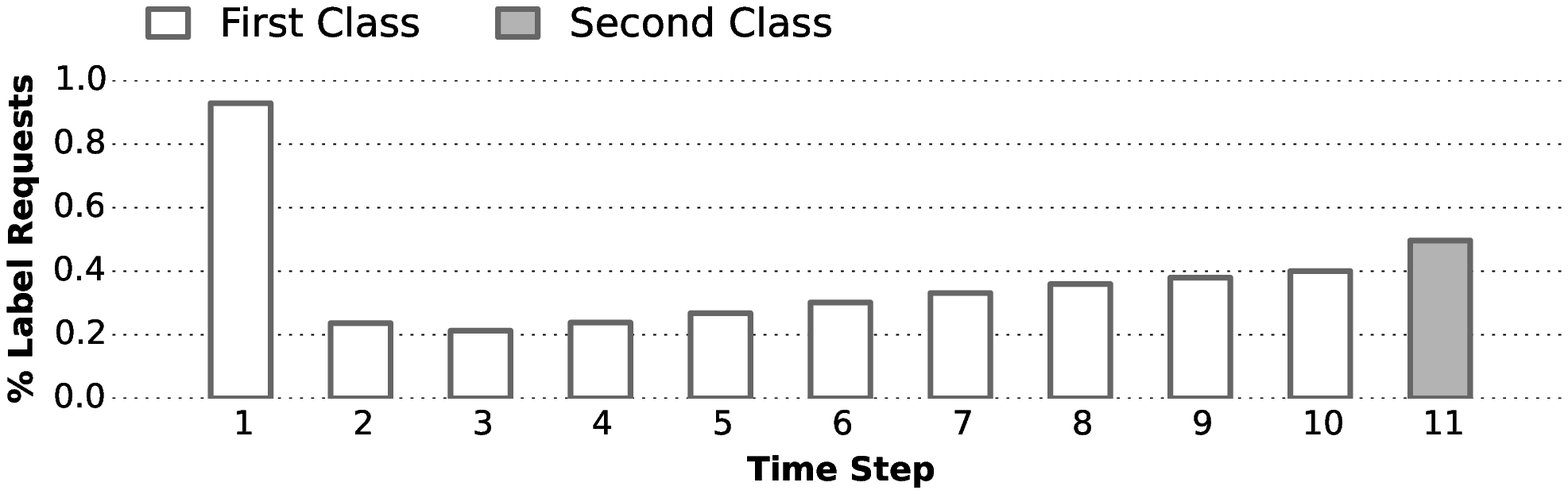}
    \caption{Switch classes after 10}
    \label{fig:class-switch-10}
  \end{subfigure}
  \caption{The trained model was further evaluated on a second task. For each episode, two random test classes were selected. (a) 5 examples from the first class are presented, followed by 1 example from the second class. (b) 10 examples from the first class are presented, followed by 1 example from the second class. The percentage of episodes in which a label request is made for each time step is shown. The difference in the percentage of label requests at time step 6, along with the similarity between the percentage of label requests when an instance of the second class is presented, suggests that the model is computing uncertainty and acting on it.}
  \label{fig:class-switch}
\end{figure}

One naive strategy that the model could use to attempt this task would be to learn a fixed time step, where it would switch from requesting labels to predicting labels. In comparison, an optimal policy would consider the model's uncertainty of the label when deciding to request a label.
To explore whether the model was using a naive strategy or effectively reasoning about its own uncertainty, we performed a second experiment using the trained model (from batch 100,000). In this experiment, two classes were selected at random and the model was presented with either 5 examples of the first class followed by 1 example of the second class, or 10 examples of the first class followed by 1 example of the second class. 
In both cases, we ran 1,000 episodes. 
For each time step, the percentage of episodes in which the model requested a label were recorded. 
If the model is using its uncertainty effectively, we should see high percentages of label requests on the first time step and on time-step when the first instance of the second class is seen. Alternatively, if uncertainty is not being used, we should see something simple such as high percentages of label requests on the first 3 time steps, followed by low percentages of label requests.
As shown in Figure~\ref{fig:class-switch}, the percentage of label requests is high on time step 1 and it is high and, notably, equal on the 6th or 11th time across the scenarios. These results are 
consistent with the model making use of its uncertainty.
The differences in label request frequency at step 6 between Fig.~\ref{fig:class-switch-5} and Fig.~\ref{fig:class-switch-10} show that the model is not following an episode independent label request schedule. The gradual increase in label requests before the 2nd class instance could suggest that the model is losing faith in its belief of the distribution of samples within a class, since a sequence of 5 or 10 images of the same class was rare at training time; further experiments are needed to confirm this.

\begin{table}[t]
  \caption{Test set classification accuracies and percentage of label requests per episode. Higher accuracy requires more labels. Reinforcement learning provides a principled approach to making that trade-off, here by specifying the reward for an incorrect answer, $R_{inc}$. With the correct choice of rewards, RL can achieve a higher prediction accuracy than supervised learning with fewer labels ($R_{inc} = -20$). The accuracy for ``RL'' includes incorrect labels for label requests. ``RL Prediction \dots'' only considers the accuracy when predictions were made. The model for ``RL'' and ``RL Prediction'' is the same model used in figure~\ref{fig:accuracy-and-requests} from step 100,000 on and in figure~\ref{fig:class-switch}, with $R_{inc}=-1$.}
  \label{accuracy-table}
  \vspace{0.1in}
  \centering
  \begin{tabular}{lrr}
    \toprule
    & \textbf{Accuracy (\%)} & \textbf{Requests (\%)} \\
    \midrule 
    Supervised & 91.0  & 100.0 \\
    RL & 75.9 & 7.2 \\
    RL Prediction & 81.8 & \textbf{7.2} \\
    RL Prediction ($R_{inc}=-5$)& 86.4 & 31.8 \\
    RL Prediction ($R_{inc}=-10$)& 89.3 & 45.6 \\
    RL Prediction ($R_{inc}=-20$)& \textbf{92.8} & 60.6 \\
    \bottomrule
  \end{tabular}
\end{table}

Finally, through the choice of rewards, we should be able to make trade-offs between high prediction accuracy with many label requests and few label requests but lower prediction accuracy. To explore this we trained several models on the task in figure~\ref{fig:task} using different values of $R_{inc}$, the reward for an incorrect prediction. We trained 100,000 episode batches and then evaluated the models on episodes containing classes from the test set. For consistency of convergence, $R_{inc}=-10$ and $R_{inc}=-20$ were trained with a batch size of 100. Table~\ref{accuracy-table} shows the results and confirms that the proposed model can smoothly make this trade-off. A supervised learning model was also evaluated, as introduced in Santoro et al.~\cite{santoro2016osl}, where the loss is the cross entropy between the predicted and true label, and the true label is always presented on the following time step. For consistency, we used our same LSTM model for this supervised task, with the modifications of a softmax on the output and not outputting the extra bit for the ``request label'' action. With the RL model we are able to achieve a higher prediction accuracy than the overall accuracy for the supervised task, while using 60.6\% of the labels at test time; the supervised task effectively requests labels 100\% of the time.

\section{Discussion \& Future Work} \label{sec:discussion-and-future-work}

We presented a method for learning an active learner via reinforcement learning. Our results demonstrate that the model can learn from only a handful of requested labels and can effectively trade off accuracy for reduced label requests, via the choice of reward. Additionally, our results suggest that the model learns to effectively reason about its uncertainty when making its decision.

An important direction for future work is to expand the experiments by increasing the complexity of the task.
Natural extensions include increasing the number of classes per episode and experimenting with more complex datasets such as ImageNet~\cite{deng2009ils}. Note that the reinforcement learning method used in our experiments was quite simplistic. By using more powerful RL strategies such as better exploration~\cite{mnih2016amd}, a separate target network~\cite{mnih2013pad}, or decomposing the action-value function~\cite{wang2015dna}, we expect our model to scale to more complex tasks. A more expressive model such as memory augmentation could also help~\cite{graves2014ntm}; though, we found that a neural turing machine with the ``least recently used'' addressing module~\cite{santoro2016osl} overfit to this task much more than an LSTM. 
Existing and future results should, where possible, be compared with prior methods for active learning.

\section{Acknowledgements} \label{sec:acknowledgements}

The authors would like to thank Adam Santoro for helpful correspondence regarding the models in Santoro et al.\cite{santoro2016osl}.

\pagebreak
\bibliography{references}

\end{document}